\documentclass{article}

\usepackage[preprint]{neurips_2025}

\workshoptitle{AI for Tabular Data}

\usepackage[utf8]{inputenc}   
\usepackage[T1]{fontenc}      
\usepackage{hyperref}         
\usepackage{url}              
\usepackage{booktabs}         
\usepackage{amsfonts}         
\usepackage{nicefrac}         
\usepackage{microtype}        
\usepackage{xcolor}           
\usepackage{amsmath}
\usepackage[pdftex]{graphicx}
\usepackage{subcaption}

\title{In Search of Grandmother Cells: Tracing Interpretable Neurons in Tabular Representations}

\author{%
  Ricardo Knauer\\
  KI-Werkstatt\\
  University of Applied Sciences Berlin\\
  Berlin, Germany
  \And
  Erik Rodner\\
  KI-Werkstatt\\
  University of Applied Sciences Berlin\\
  Berlin, Germany
}

\begin{document}

\maketitle

\begin{abstract}

Foundation models are powerful yet often opaque in their decision-making. A topic of continued interest in both neuroscience and artificial intelligence is whether some neurons behave like ``grandmother cells'', \textit{i.e.}, neurons that are inherently interpretable because they exclusively respond to single concepts. In this work, we propose two information-theoretic measures that quantify the neuronal saliency and selectivity for single concepts. We apply these metrics to the representations of TabPFN, a tabular foundation model, and perform a simple search across neuron-concept pairs to find the most salient and selective pair. Our analysis provides the first evidence that some neurons in such models show moderate, statistically significant saliency and selectivity for high-level concepts. These findings suggest that interpretable neurons can emerge naturally and that they can, in some cases, be identified without resorting to more complex interpretability techniques.

\end{abstract}

\section{Introduction}

The rise of foundation models has unlocked powerful capabilities, but their decision-making processes often remain opaque \citep{bommasani2022opportunitiesrisks,longo2024explainable}. A recurring theme in the fields of neuroscience \citep{gross2002genealogy,quiroga2005invariant} and artificial intelligence \citep{arora2018linear,olah2020zoom} is whether some neurons act like ``grandmother cells'', \textit{i.e.}, neurons that exclusively respond to single high-level features or concepts. Understanding the nature of neuronal representations can facilitate their interpretation and foster trust among users, especially in high-stake domains such as the clinical setting \citep{adler2022deutsche,eu2024aiact}. In recent years, many advanced techniques have been developed within the subfield of mechanistic interpretability to gain deeper insights into the inner workings of foundation models, \textit{e.g.}, by identifying interpretable units of computation \citep{ferrando2024primerinnerworkings,sharkey2025openproblems}. In our work, we examine whether we actually always need sophisticated methods to find interpretable units. 

Our contributions are as follows:

\begin{itemize}
    \item We propose \textbf{two information-theoretic measures of neuronal saliency and selectivity for single concepts} \citep{hewitt2019designing,kandel2021principles,simonyan2014deepinsidecnns}. We use hypothesis testing to find neurons that are significantly salient and selective for single concepts and analytically derive the null distributions for our metrics (Sect.~\ref{sec:formal}).
    \item We apply our metrics to the representations of TabPFN \citep{hollmann2025accurate}, a tabular foundation model, and offer the first evidence that \textbf{some neurons in tabular foundation models exhibit moderate, statistically significant saliency and selectivity for high-level concepts} (Sect.~\ref{sec:experiments}). Our findings suggest that interpretable neurons can emerge naturally and that they can sometimes be uncovered without relying on more complex interpretability techniques.
\end{itemize}

\section{Related work}

Whether individual neurons correspond to single concepts, \textit{i.e.}, behave like ``grandmother cells'', remains a fundamental question in both neuroscience \citep{gross2002genealogy,quiroga2005invariant} and artificial intelligence \citep{arora2018linear,olah2020zoom}. Formally, a neural network cannot represent concepts using orthogonal bases if the number of independent concepts is greater than the number of neurons in the network. As a result, concepts are often encoded by multiple neurons (distributed representation) and neurons often encode multiple concepts (superposition) \citep{arora2018linear,olah2020zoom}. Although there are interpretability techniques to localize and disentangle such representations, \textit{e.g.}, sparse probes \citep{bertsimas2021sparse,gurnee2023finding} and sparse autoencoders \citep{gao2025scaling,lieberum2024gemma,templeton2024scaling}, they usually require training auxiliary models and incur high computational costs. This raises the question whether simpler methods could sometimes suffice to identify interpretable neurons.

In the next section, we propose to quantify the neuronal saliency and selectivity for single concepts \citep{hewitt2019designing,kandel2021principles,simonyan2014deepinsidecnns} using information theory \citep{dayan2005theoretical,li2024can,shannon1948mathematical}, without requiring the training of auxiliary models.

\section{Quantifying neuronal interpretability} \label{sec:formal}

In this section, we define our two information-theoretic measures of neuronal saliency and selectivity for single concepts, and analytically derive their null distributions for hypothesis testing.

\subsection{Metric definitions} \label{sec:definitions}

Let $\mathbf{A} \in \mathbb{R}^{M \times N}$ be the neuronal activations for $M \in \mathbb{N_+}$ samples and $N \in \mathbb{N_+}$ neurons. The network structure is not relevant for the subsequent definitions and therefore abstracted away. Moreover, let $\mathbf{B} \in \{0, 1\}^{M \times C} $ represent the concept labels for the $M$ samples and $C \in \mathbb{N_+}$ high-level, interpretable features, \textit{i.e.}, concepts. We measure the association between the activations $\mathbf{a}_i \in \mathbb{R}^M$ of neuron $i$ and the concept labels $\mathbf{b}_j \in \{0, 1\}^M$ of concept $j$ with the empirical mutual information $\hat{I}(\mathbf{a}_i, \mathbf{b}_j) \in \mathbb{R}_{+ \cup \{0\}}$ \citep{dayan2005theoretical,li2024can,shannon1948mathematical}. We further denote by
\begin{align}
\hat{p}_{tail}(\mathbf{a}_i, \mathbf{b}_j) = \frac{1}{N} \sum_{n=1}^{N} [\hat{I}(\mathbf{a}_n, \mathbf{b}_j) \ge \hat{I}(\mathbf{a}_i, \mathbf{b}_j)], \quad \frac{1}{N} \leq \hat{p}_{tail}(\mathbf{a}_i, \mathbf{b}_j) \leq 1
\label{eq:p_tail}
\end{align}
the empirical upper-tail probability of $\hat{I}(\mathbf{a}_i, \mathbf{b}_j)$, \textit{i.e.}, the fraction of neurons whose empirical mutual information to concept $j$ is greater than or equal to that of neuron $i$, where $[\cdot]$ are the Iverson brackets.

We quantify the saliency \citep{kandel2021principles, simonyan2014deepinsidecnns} for neuron $i$ with respect to concept $j$ using the information content or surprisal of observing $\hat{p}_{tail}(\mathbf{a}_i, \mathbf{b}_j)$. This captures the intuition that unexpected observations indicate more pronounced neuron–concept relationships:
\begin{align}
\text{surprisal}(\mathbf{a}_i, \mathbf{b}_j) = -\log (\hat{p}_{tail}(\mathbf{a}_i, \mathbf{b}_j)), \quad 0 \leq \text{surprisal}(\mathbf{a}_i, \mathbf{b}_j) \leq \log(N) \quad.
\label{eq:surprisal}
\end{align}
Then, we define the selectivity \citep{hewitt2019designing,kandel2021principles} as:
\begin{align}
\text{selectivity}(\mathbf{a}_i, \mathbf{b}_j) = \frac{\text{surprisal}(\mathbf{a}_i, \mathbf{b}_j)}{\sum_{c=1}^{C} \text{surprisal}(\mathbf{a}_i, \mathbf{b}_c)}, \quad 0 \leq \text{selectivity}(\mathbf{a}_i, \mathbf{b}_j) \leq 1 \quad.
\label{eq:selectivity}
\end{align}
Please note that smaller concept set sizes tend to yield higher selectivity values due to the normalization, and that the selectivity only accounts for concepts $c\in \{1, ..., C\}$. In the following, we derive the upper-tail p-values for $\text{surprisal}(\mathbf{a}_i, \mathbf{b}_j)$ and $\text{selectivity}(\mathbf{a}_i, \mathbf{b}_j)$ in isolation and in combination.

\subsection{Significance testing} \label{sec:significance}

Our null hypothesis $H_0$ is that neuron-concept associations are exchangeable, \textit{i.e.}, no neuron is preferentially associated with any concept and no concept is preferentially associated with any neuron. We assume that the mutual information scores are independent and identically distributed (i.i.d.) across neurons for concept $j$ and that the surprisals are i.i.d. across concepts for neuron $i$. As $N\to\infty$, this implies that $\hat{p}_{tail}(\cdot, \mathbf{b}_j) \Rightarrow \text{Uniform}(0, 1)$ and $\text{surprisal}(\cdot, \mathbf{b}_j) \Rightarrow \text{Exponential}(1) = \text{Gamma}(1, 1)$. Normalizing $C$ i.i.d. \text{Gamma}(1, 1) random variables yields a flat $\text{Dirichlet}(1, ..., 1)$ distribution with $\text{Beta}(1, C - 1)$ marginals, so that the selectivity for neuron $i$ with respect to each concept $j\in \{1, ..., C\}$ is $\text{Beta}(1, C - 1)$-distributed.

The upper-tail p-value of observing $\text{surprisal}(\mathbf{a}_i, \mathbf{b}_j)$ is:
\begin{align}
P(\text{surprisal}(\cdot, \mathbf{b}_j) \geq \text{surprisal}(\mathbf{a}_i, \mathbf{b}_j) \,|\, H_0) = \hat{p}_{tail}(\mathbf{a}_i, \mathbf{b}_j) \quad.
\label{eq:p_surprisal}
\end{align}
The upper-tail p-value of observing $\text{selectivity}(\mathbf{a}_i, \mathbf{b}_j)$ is \citep{olver2010nist}:
\begin{align}
P(\text{selectivity}(\mathbf{a}_i, \cdot) \geq \text{selectivity}(\mathbf{a}_i, \mathbf{b}_j) \,|\, H_0) = (1 - \text{selectivity}(\mathbf{a}_i, \mathbf{b}_j))^{C - 1} \quad.
\label{eq:p_selectivity}
\end{align}
To enforce a conservative family-wise error rate control, we combine both p-values with a Bonferroni correction that adjusts the minimum p-value for the number of tests \citep{nikolitsa2025metacp}:
\begin{align}
p_{comb}(\mathbf{a}_i, \mathbf{b}_j) = \min(2NC \cdot \min(\hat{p}_{tail}(\mathbf{a}_i, \mathbf{b}_j), (1 - \text{selectivity}(\mathbf{a}_i, \mathbf{b}_j))^{C - 1}), 1) \quad.
\label{eq:p_comb}
\end{align}

In the next section, we use our definitions of $\text{surprisal}(\mathbf{a}_i, \mathbf{b}_j)$ (Eq.~\eqref{eq:surprisal}) and $\text{selectivity}(\mathbf{a}_i, \mathbf{b}_j)$  (Eq.~\eqref{eq:selectivity}) as well as $p_{comb}(\mathbf{a}_i, \mathbf{b}_j)$ (Eq.~\eqref{eq:p_comb}) to empirically test whether we can use our metrics to find interpretable neurons in foundation models.

\section{Experiments} \label{sec:experiments}

In this section, we use the surprisal and selectivity (Sect.~\ref{sec:definitions}) as well as their combined p-value (Sect.~\ref{sec:significance}) to evaluate whether we can use our measures to uncover interpretable neurons in foundation models. To this end, we apply them to the representations of TabPFN \citep{hollmann2025accurate}, a tabular foundation model, and provide the first evidence that some neurons in such models show moderate, statistically significant saliency and selectivity for high-level concepts. These findings suggest that interpretable neurons can emerge naturally and that, in some cases, they can be identified without relying on more complex interpretability techniques.

\subsection{Experimental setup} \label{sec:setup}

We focused our experiments on a setting with clearly defined concept labels. Specifically, we treated diagnostic codes from the International Classification of Diseases (ICD) as concepts organized at three hierarchical levels (low, mid, and high) and considered them for four tabular prediction tasks at emergency department (ED) triage: inhospital mortality, intensive care unit (ICU) transfer within 12h, critical outcome, and hospitalization prediction \citep{xie2022benchmarking}. Please refer to Appendix~\ref{sec:appendix_datasets} for details on the datasets and preprocessing.

We employed TabPFN 2.0.9 \citep{hollmann2025accurate} as our foundation model due to its strong reported performance \citep{erickson2025tabarena}. We used TabPFN without tuning or ensembling and evaluated its discriminative performance with the test set area under the receiver operating characteristic curve (AUC). We then extracted 192-dimensional embeddings from the 12 encoder layers, computed the surprisal and selectivity for all neuron-concept pairs, and treated the knee point on the surprisal-selectivity Pareto front as our most salient and selective neuron-concept pair. To identify the knee point, we maximized the sum of the min-max scaled surprisal and selectivity scores. We additionally computed $p_{comb}(\mathbf{a}_i, \mathbf{b}_j)$ (Eq.~\eqref{eq:p_comb}) for the knee point and selected the top-3 input features for the knee point activations based on their mutual information when the concept was present. As baselines, we employed sparse probes based on SHAP values \citep{covert2021explaining,lundberg2017unified} and optimal probing \citep{bertsimas2021sparse,gurnee2023finding}. Please refer to Appendix~\ref{sec:appendix_baselines} for details.

\subsection{Experimental results}

TabPFN's test set AUC ranged from acceptable (0.79) to excellent (0.88), matching the best machine learning baselines in \citet{xie2022benchmarking}. Most neurons in TabPFN were neither salient nor selective for specific ICD codes (Fig.~\ref{fig:task_1_3} and~\ref{fig:task_0_2} in Appendix~\ref{sec:appendix_results}). Nevertheless, some neuron-concept pairs could reach maximum surprisal values of $\log(N) = 7.74$ or moderate selectivity values of up to $0.73$. Sparse probes based on SHAP values and optimal probing were Pareto-dominated by our search procedure (Fig.~\ref{fig:task_1_3} and~\ref{fig:task_0_2} in Appendix~\ref{sec:appendix_results}).

In Table~\ref{tab:results}, we report the most salient and selective neuron-concept pairs for each of our four tasks, \textit{i.e.}, the knee points on the surprisal-selectivity Pareto fronts (Fig.~\ref{fig:task_1_3} and~\ref{fig:task_0_2} in Appendix~\ref{sec:appendix_results}). Most knee points and indeed many points on the Pareto fronts corresponded to ICD code V (transport accidents). For instance, TabPFN neuron 49 in layer 8 may potentially encode high triage acuity (emergency severity index), hypoxemia (low oxygen saturation), and other features as signs of a traumatic chest injury due to a transport accident, which typically requires hospitalization (Table~\ref{tab:results}). We found knee points across initial, middle, and final layers in TabPFN, suggesting that concepts may not be preferentially represented at certain layers. The surprisal and selectivity values for the knee points reached statistical significance on half of our tasks ($p<0.05$).

Overall, we provide the first evidence that interpretable neurons can emerge naturally in tabular foundation models and that they can be uncovered, at least in some cases, with a simple search across neuron-concept pairs.

\begin{table}[t]
  \footnotesize
  \caption{Knee points on the surprisal-selectivity Pareto fronts. ICD code Q corresponds to congenital malformations, deformations, and chromosomal abnormalities, ICD code V to transport accidents.}
  \label{tab:results}
  \centering
  \begin{tabular}{p{1.4cm} p{0.8cm}p{0.5cm}p{0.9cm} p{3.4cm} p{1.0cm}p{1.2cm}p{1.3cm}}
    \toprule
    Prediction target & Neuron & Layer & Concept & Top-3 features & Surprisal & Selectivity & p-value \\
    \midrule
    \multicolumn{7}{l}{\bfseries\boldmath Statistically significant results ($p < 0.05$)} \\
    \addlinespace[1ex]
    ICU transfer within 12h & $113$ & $12$ & Q & ICU admissions in the past year, heart rate, hypothyroidism & $6.13$ & $0.73$ & $2.1 \cdot 10^{-6}$ \\
    \addlinespace[1ex]
    Hospitali-zation & $49$ & $8$ & V & Hospitalizations in the past three months, oxygen saturation, emergency severity index & $5.80$ & $0.62$ & $5.7 \cdot 10^{-3}$ \\
    \midrule
    \multicolumn{7}{l}{\bfseries\boldmath Statistically insignificant results ($p \geq 0.05$)} \\
    \addlinespace[1ex]
    Inhospital mortality & $78$ & $9$ & V & ED visits in the past month, respiration rate, dementia & $5.80$ & $0.53$ & $0.89$ \\
    \addlinespace[1ex]
    Critical outcome & $62$ & $1$ & V & ED visits in the past three months, pain scale, cardiac arrhythmia & $7.74$ & $0.30$ & $1.00$ \\
    \bottomrule
  \end{tabular}
\end{table}

\section{Limitations}

Although our findings provide valuable insights into the inner workings of foundation models, they should be interpreted in light of several limitations. First, our significance tests are asymptotic and assume i.i.d. mutual information and surprisal scores, which does not account for variations in empirical distributions (\textit{e.g.}, due to data scarcity) nor for structural dependencies in the network architecture (\textit{e.g.}, layers) or concept label space (\textit{e.g.}, hierarchies). Second, the surprisal and selectivity scores did not reach statistical significance in all tasks, which may reflect that inpatient mortality depends on multiple interacting factors rather than single ICD codes, whereas ICU transfer or hospitalization are more strongly driven by deterministic referral policies. What is more, the empirical mutual information ranking (Eq.~\eqref{eq:p_tail}) reduces the statistical power, so that the statistical significance is effectively determined by the selectivity (Eq.~\eqref{eq:p_comb}). Third, the selectivity values remained at or below $0.73$, meaning that we were not able to identify ``grandmother cells'' that (almost) exclusively respond to single concepts.

\section{Conclusion}

Interpreting the decision-making processes of foundation models remains an open challenge. In our work, we propose an information-theoretic testing framework that allows us to identify neurons that are both salient and selective for single concepts. Applied to the representations of TabPFN, a tabular foundation model, our framework offers the first evidence that some neurons in such models show moderate, statistically significant saliency and selectivity for high-level concepts. We are confident that our framework serves as a useful tool for mechanistic interpretability researchers and practitioners to analyze concept representations in foundation models.

\begin{ack}
This research was funded by the Bundesministerium für Bildung und Forschung (BMBF, project number: 16DHBKI071) and the Deutsche Forschungsgemeinschaft (DFG, German Research Foundation, project number: 528483508). The authors declare no competing interests.

\end{ack}

\bibliography{bibliography}

@article{adler2022deutsche,
  title={Deutsche {N}ormungsroadmap {K}{\"u}nstliche {I}ntelligenz},
  author={Adler, Rasmus and Bunte, Andreas and Burton, Simon and Gro{\ss}mann, J{\"u}rgen and Jaschke, Alexander and Kleen, Philip and Lorenz, Jeanette Miriam and Ma, Jackie and Markert, Karla and Mee{\ss}, Henri and others},
  year={2022},
  publisher={DIN}
}

@article{arora2018linear,
  title={Linear algebraic structure of word senses, with applications to polysemy},
  author={Arora, Sanjeev and Li, Yuanzhi and Liang, Yingyu and Ma, Tengyu and Risteski, Andrej},
  journal={Transactions of the Association for Computational Linguistics},
  volume={6},
  pages={483--495},
  year={2018},
  publisher={MIT Press One Rogers Street, Cambridge, MA 02142-1209, USA journals-info~…}
}

@misc{bommasani2022opportunitiesrisks,
      title={On the Opportunities and Risks of Foundation Models}, 
      author={Rishi Bommasani and Drew A. Hudson and Ehsan Adeli and Russ Altman and Simran Arora and Sydney von Arx and Michael S. Bernstein and Jeannette Bohg and Antoine Bosselut and Emma Brunskill and Erik Brynjolfsson and Shyamal Buch and Dallas Card and Rodrigo Castellon and Niladri Chatterji and Annie Chen and Kathleen Creel and Jared Quincy Davis and Dora Demszky and Chris Donahue and Moussa Doumbouya and Esin Durmus and Stefano Ermon and John Etchemendy and Kawin Ethayarajh and Li Fei-Fei and Chelsea Finn and Trevor Gale and Lauren Gillespie and Karan Goel and Noah Goodman and Shelby Grossman and Neel Guha and Tatsunori Hashimoto and Peter Henderson and John Hewitt and Daniel E. Ho and Jenny Hong and Kyle Hsu and Jing Huang and Thomas Icard and Saahil Jain and Dan Jurafsky and Pratyusha Kalluri and Siddharth Karamcheti and Geoff Keeling and Fereshte Khani and Omar Khattab and Pang Wei Koh and Mark Krass and Ranjay Krishna and Rohith Kuditipudi and Ananya Kumar and Faisal Ladhak and Mina Lee and Tony Lee and Jure Leskovec and Isabelle Levent and Xiang Lisa Li and Xuechen Li and Tengyu Ma and Ali Malik and Christopher D. Manning and Suvir Mirchandani and Eric Mitchell and Zanele Munyikwa and Suraj Nair and Avanika Narayan and Deepak Narayanan and Ben Newman and Allen Nie and Juan Carlos Niebles and Hamed Nilforoshan and Julian Nyarko and Giray Ogut and Laurel Orr and Isabel Papadimitriou and Joon Sung Park and Chris Piech and Eva Portelance and Christopher Potts and Aditi Raghunathan and Rob Reich and Hongyu Ren and Frieda Rong and Yusuf Roohani and Camilo Ruiz and Jack Ryan and Christopher Ré and Dorsa Sadigh and Shiori Sagawa and Keshav Santhanam and Andy Shih and Krishnan Srinivasan and Alex Tamkin and Rohan Taori and Armin W. Thomas and Florian Tramèr and Rose E. Wang and William Wang and Bohan Wu and Jiajun Wu and Yuhuai Wu and Sang Michael Xie and Michihiro Yasunaga and Jiaxuan You and Matei Zaharia and Michael Zhang and Tianyi Zhang and Xikun Zhang and Yuhui Zhang and Lucia Zheng and Kaitlyn Zhou and Percy Liang},
      year={2022},
      eprint={2108.07258},
      archivePrefix={arXiv},
      primaryClass={cs.LG},
      url={https://arxiv.org/abs/2108.07258}, 
}

@article{bertsimas2021sparse,
  title={Sparse classification: a scalable discrete optimization perspective},
  author={Bertsimas, Dimitris and Pauphilet, Jean and Van Parys, Bart},
  journal={Machine Learning},
  volume={110},
  number={11},
  pages={3177--3209},
  year={2021},
  publisher={Springer}
}

@article{covert2021explaining,
  title={Explaining by removing: A unified framework for model explanation},
  author={Covert, Ian and Lundberg, Scott and Lee, Su-In},
  journal={Journal of Machine Learning Research},
  volume={22},
  number={209},
  pages={1--90},
  year={2021}
}

@book{dayan2005theoretical,
  title={Theoretical neuroscience: computational and mathematical modeling of neural systems},
  author={Dayan, Peter and Abbott, Laurence F},
  year={2005},
  publisher={MIT press}
}

@misc{erickson2025tabarena,
      title={{TabArena}: A Living Benchmark for Machine Learning on Tabular Data}, 
      author={Nick Erickson and Lennart Purucker and Andrej Tschalzev and David Holzmüller and Prateek Mutalik Desai and David Salinas and Frank Hutter},
      year={2025},
      eprint={2506.16791},
      archivePrefix={arXiv},
      primaryClass={cs.LG},
      url={https://arxiv.org/abs/2506.16791}, 
}

@article{erion2021improving,
  title={Improving performance of deep learning models with axiomatic attribution priors and expected gradients},
  author={Erion, Gabriel and Janizek, Joseph D and Sturmfels, Pascal and Lundberg, Scott M and Lee, Su-In},
  journal={Nature machine intelligence},
  volume={3},
  number={7},
  pages={620--631},
  year={2021},
  publisher={Nature Publishing Group UK London}
}

@misc{eu2024aiact,
  title="{Regulation (EU) 2024/1689 of the European Parliament and of the Council of 13 June 2024 laying down harmonised rules on artificial intelligence and amending Regulations (EC) No 300/2008, (EU) No 167/2013, (EU) No 168/2013, (EU) 2018/858, (EU) 2018/1139 and (EU) 2019/2144 and Directives 2014/90/EU, (EU) 2016/797 and (EU) 2020/1828 (Artificial Intelligence Act)}",
  author="{European Parliament and Council of the European Union}",
  year={2024},
  journaltitle="{Official Journal of the European Union, L series}",
  url={https://eur-lex.europa.eu/eli/reg/2024/1689/oj}
}

@misc{ferrando2024primerinnerworkings,
      title={A Primer on the Inner Workings of Transformer-based Language Models}, 
      author={Javier Ferrando and Gabriele Sarti and Arianna Bisazza and Marta R. Costa-jussà},
      year={2024},
      eprint={2405.00208},
      archivePrefix={arXiv},
      primaryClass={cs.CL},
      url={https://arxiv.org/abs/2405.00208}, 
}

@inproceedings{gao2025scaling,
  title={Scaling and evaluating sparse autoencoders},
  author={Gao, Leo and la Tour, Tom Dupr{\'e} and Tillman, Henk and Goh, Gabriel and Troll, Rajan and Radford, Alec and Sutskever, Ilya and Leike, Jan and Wu, Jeffrey},
  booktitle={Proceedings of the 13th International Conference on Learning Representations (ICLR)},
  year={2025},
  address={Singapore},
  publisher={International Conference on Learning Representations (ICLR)},
  url={https://openreview.net/forum?id=tcsZt9ZNKD}
}

@article{gross2002genealogy,
  title={Genealogy of the “grandmother cell”},
  author={Gross, Charles G},
  journal={The Neuroscientist},
  volume={8},
  number={5},
  pages={512--518},
  year={2002},
  publisher={Sage Publications Sage CA: Thousand Oaks, CA}
}

@article{gurnee2023finding,
  title={Finding neurons in a haystack: Case studies with sparse probing},
  author={Gurnee, Wes and Nanda, Neel and Pauly, Matthew and Harvey, Katherine and Troitskii, Dmitrii and Bertsimas, Dimitris},
  journal={Transactions on Machine Learning Research},
  year={2023}
}

@inproceedings{hewitt2019designing,
    title={Designing and Interpreting Probes with Control Tasks},
    author={Hewitt, John  and Liang, Percy},
    booktitle={Proceedings of the 2019 Conference on Empirical Methods in Natural Language Processing and the 9th International Joint Conference on Natural Language Processing (EMNLP-IJCNLP)},
    year={2019},
    address={Hong Kong, China},
    publisher={Association for Computational Linguistics (ACL)},
    url={https://aclanthology.org/D19-1275/},
    doi={10.18653/v1/D19-1275},
    pages={2733--2743}
}

@article{hollmann2025accurate,
  title={Accurate predictions on small data with a tabular foundation model},
  author={Hollmann, Noah and M{\"u}ller, Samuel and Purucker, Lennart and Krishnakumar, Arjun and K{\"o}rfer, Max and Hoo, Shi Bin and Schirrmeister, Robin Tibor and Hutter, Frank},
  journal={Nature},
  volume={637},
  number={8045},
  pages={319--326},
  year={2025},
  publisher={Nature Publishing Group UK London}
}

@book{kandel2021principles,
  title={Principles of neural science},
  author={Kandel, Eric R and Koester, John D and Mack, Sarah H and Siegelbaum, Steven A},
  volume={6},
  year={2021},
  publisher={McGraw Hill}
}

@article{li2024can,
  title={{CAN}: Concept-Aligned Neurons for Visual Comparison of Deep Neural Network Models},
  author={Li, Mingwei and Jeong, Sangwon and Liu, Shusen and Berger, Matthew},
  journal={Computer Graphics Forum},
  volume={43},
  number={3},
  pages={e15085},
  year={2024},
  publisher={Wiley Online Library}
}

@inproceedings{lieberum2024gemma,
    title={Gemma {S}cope: Open Sparse Autoencoders Everywhere All At Once on {G}emma 2},
    author={Lieberum, Tom  and
      Rajamanoharan, Senthooran  and
      Conmy, Arthur  and
      Smith, Lewis  and
      Sonnerat, Nicolas  and
      Varma, Vikrant  and
      Kramar, Janos  and
      Dragan, Anca  and
      Shah, Rohin  and
      Nanda, Neel},
    booktitle={Proceedings of the 7th BlackboxNLP Workshop: Analyzing and Interpreting Neural Networks for NLP},
    year={2024},
    address={Miami, USA},
    publisher={Association for Computational Linguistics (ACL)},
    url={https://aclanthology.org/2024.blackboxnlp-1.19/},
    doi={10.18653/v1/2024.blackboxnlp-1.19},
    pages={278--300}
}

@article{longo2024explainable,
  title={Explainable Artificial Intelligence ({XAI}) 2.0: A manifesto of open challenges and interdisciplinary research directions},
  author={Longo, Luca and Brcic, Mario and Cabitza, Federico and Choi, Jaesik and Confalonieri, Roberto and Del Ser, Javier and Guidotti, Riccardo and Hayashi, Yoichi and Herrera, Francisco and Holzinger, Andreas and others},
  journal={Information Fusion},
  volume={106},
  pages={102301},
  year={2024},
  publisher={Elsevier}
}

@article{lundberg2017unified,
  title={A unified approach to interpreting model predictions},
  author={Lundberg, Scott M and Lee, Su-In},
  journal={Advances in neural information processing systems},
  volume={30},
  year={2017}
}

@article{nikolitsa2025metacp,
  title={metacp: a versatile software package for combining dependent or independent p-values},
  author={Nikolitsa, Evgenia K and Kontou, Panagiota I and Bagos, Pantelis G},
  journal={BMC Bioinformatics},
  volume={26},
  number={1},
  pages={109},
  year={2025},
  publisher={Springer}
}

@article{olah2020zoom,
  title={Zoom in: An introduction to circuits},
  author={Olah, Chris and Cammarata, Nick and Schubert, Ludwig and Goh, Gabriel and Petrov, Michael and Carter, Shan},
  journal={Distill},
  volume={5},
  number={3},
  pages={e00024--001},
  year={2020}
}

@book{olver2010nist,
  title={NIST handbook of mathematical functions},
  author={Olver, Frank W J and Lozier, Daniel W and Boisvert, Ronald F and Clark, Charles W},
  year={2010},
  publisher={Cambridge University Press}
}

@article{quiroga2005invariant,
  title={Invariant visual representation by single neurons in the human brain},
  author={Quiroga, R Quian and Reddy, Leila and Kreiman, Gabriel and Koch, Christof and Fried, Itzhak},
  journal={Nature},
  volume={435},
  number={7045},
  pages={1102--1107},
  year={2005},
  publisher={Nature Publishing Group UK London}
}

@article{shannon1948mathematical,
  title={A mathematical theory of communication},
  author={Shannon, Claude E},
  journal={The Bell System Technical Journal},
  volume={27},
  number={3},
  pages={379--423},
  year={1948},
  publisher={Nokia Bell Labs}
}

@misc{sharkey2025openproblems,
      title={Open Problems in Mechanistic Interpretability}, 
      author={Lee Sharkey and Bilal Chughtai and Joshua Batson and Jack Lindsey and Jeff Wu and Lucius Bushnaq and Nicholas Goldowsky-Dill and Stefan Heimersheim and Alejandro Ortega and Joseph Bloom and Stella Biderman and Adria Garriga-Alonso and Arthur Conmy and Neel Nanda and Jessica Rumbelow and Martin Wattenberg and Nandi Schoots and Joseph Miller and Eric J. Michaud and Stephen Casper and Max Tegmark and William Saunders and David Bau and Eric Todd and Atticus Geiger and Mor Geva and Jesse Hoogland and Daniel Murfet and Tom McGrath},
      year={2025},
      eprint={2501.16496},
      archivePrefix={arXiv},
      primaryClass={cs.LG},
      url={https://arxiv.org/abs/2501.16496}, 
}

@inproceedings{shrikumar2017learning,
  title={Learning important features through propagating activation differences},
  author={Shrikumar, Avanti and Greenside, Peyton and Kundaje, Anshul},
  booktitle={International Conference on Machine Learning (ICML)},
  pages={3145--3153},
  year={2017},
  organization={PMLR}
}

@misc{simonyan2014deepinsidecnns,
      title={Deep Inside Convolutional Networks: Visualising Image Classification Models and Saliency Maps}, 
      author={Karen Simonyan and Andrea Vedaldi and Andrew Zisserman},
      year={2014},
      eprint={1312.6034},
      archivePrefix={arXiv},
      primaryClass={cs.CV},
      url={https://arxiv.org/abs/1312.6034}, 
}

@inproceedings{sundararajan2017axiomatic,
  title={Axiomatic attribution for deep networks},
  author={Sundararajan, Mukund and Taly, Ankur and Yan, Qiqi},
  booktitle={International Conference on Machine Learning (ICML)},
  pages={3319--3328},
  year={2017},
  organization={PMLR}
}

@article{templeton2024scaling,
       title={Scaling Monosemanticity: Extracting Interpretable Features from {C}laude 3 {S}onnet},
       author={Templeton, Adly and Conerly, Tom and Marcus, Jonathan and Lindsey, Jack and Bricken, Trenton and Chen, Brian and Pearce, Adam and Citro, Craig and Ameisen, Emmanuel and Jones, Andy and Cunningham, Hoagy and Turner, Nicholas L and McDougall, Callum and MacDiarmid, Monte and Freeman, C. Daniel and Sumers, Theodore R. and Rees, Edward and Batson, Joshua and Jermyn, Adam and Carter, Shan and Olah, Chris and Henighan, Tom},
       year={2024},
       journal={Transformer Circuits Thread},
       url={https://transformer-circuits.pub/2024/scaling-monosemanticity/index.html}
    }

@article{xie2022benchmarking,
  title={Benchmarking emergency department prediction models with machine learning and public electronic health records},
  author={Xie, Feng and Zhou, Jun and Lee, Jin Wee and Tan, Mingrui and Li, Siqi and Rajnthern, Logasan S/O and Chee, Marcel Lucas and Chakraborty, Bibhas and Wong, An-Kwok Ian and Dagan, Alon and others},
  journal={Scientific Data},
  volume={9},
  number={1},
  pages={658},
  year={2022},
  publisher={Nature Publishing Group UK London}
}
\bibliographystyle{apalike}

\clearpage
\appendix

\section{Datasets and preprocessing} \label{sec:appendix_datasets}

We employed the MIMIC-IV-ED dataset to tackle four tabular prediction tasks at ED triage as described by \citet{xie2022benchmarking}: inhospital mortality prediction, ICU transfer within 12h prediction, critical outcome prediction (defined as either inhospital mortality or ICU transfer within 12h prediction), and hospitalization prediction, each treated as a binary classification problem. The data, collected at the Beth Israel Deaconess Medical Center between 2011 and 2019, comprised 64 features spanning patient history, demographics, vital signs, and chief complaints. The training set contained 353,150 ED episodes (samples) from 182,588 unique patients, the test set 88,287 ED episodes from 65,169 unique patients \citep{xie2022benchmarking}. We included rows with available ICD codes and extracted concepts at three hierarchical levels:

\begin{itemize}
\item Low-level (\textit{e.g.}, R570 = cardiogenic shock) with 5684 concepts,
\item Mid-level (\textit{e.g.} R57 = shock) with 1239 concepts,
\item High-level (\textit{e.g.}, R = signs and symptoms) with 24 concepts.
\end{itemize}

For all tasks except hospitalization prediction, the minority class constituted $\leq 7\%$ of the training samples. To address the class imbalance, we undersampled the majority classes in the training sets for inhospital mortality, ICU transfer, and critical outcome prediction. For hospitalization prediction, where the classes were relatively balanced, we undersampled both classes to 5000 samples due to the context length limitations of our tabular foundation model (Sect.~\ref{sec:setup}). This yielded 512, 2712, 2948, and 10000 training samples for mortality, ICU transfer, critical outcome, and hospitalization prediction, respectively.

\section{Baselines}
\label{sec:appendix_baselines}

To contextualize the performance of our search procedure, we compared it against two sparse probing baselines. Both approaches used linear probes to predict the concept labels from the neuronal activations. The first baseline employed a local game-theoretic neuron selection method based on SHAP values \citep{covert2021explaining,lundberg2017unified}, while the second used a globally optimal neuron selection strategy \citep{bertsimas2021sparse,gurnee2023finding}:

\begin{itemize}
    \item In the SHAP-based approach, we trained one multivariate logistic regression classifier per concept and selected the neuron with the largest mean absolute interventional SHAP value \citep{covert2021explaining,lundberg2017unified}. In this case, the method is equivalent to common gradient-based explanations when mean-centering is applied, including gradient $\times$ input \citep{shrikumar2017learning}, DeepLIFT \citep{shrikumar2017learning}, integrated gradients \citep{sundararajan2017axiomatic}, and expected gradients \citep{erion2021improving}.
    \item In the optimal probing approach, we trained $N$ univariate logistic regression classifiers per concept and selected the neuron whose logistic model achieved the largest log-likelihood. This exhaustive enumeration is equivalent to using optimization solvers \citep{bertsimas2021sparse,gurnee2023finding}, but is computationally more efficient for single-neuron selection.
\end{itemize}

For each baseline, this resulted in $C$ selected neurons, one per concept.

\newpage
\section{Pareto front visualization} \label{sec:appendix_results}

\begin{figure}[h]
  \centering
  \begin{subfigure}{0.49\textwidth}
    \includegraphics[width=\textwidth]{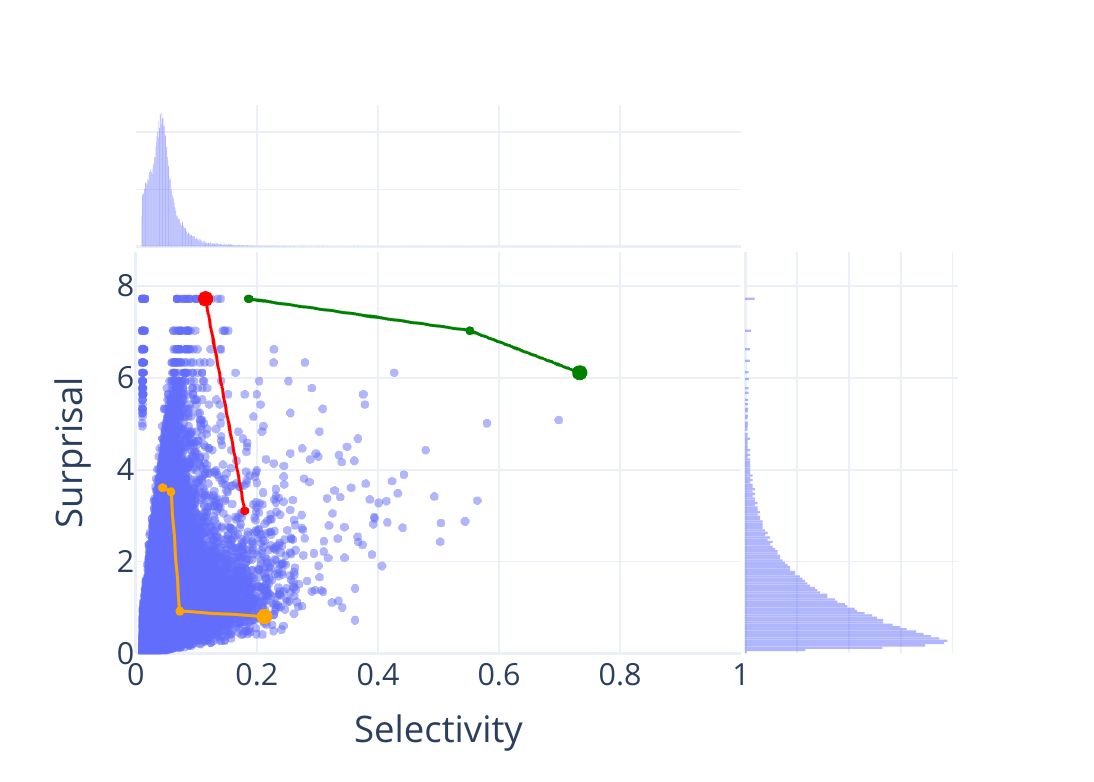}
    \caption{ICU transfer within 12h prediction.}
    \label{fig:task_1}
  \end{subfigure}
  \begin{subfigure}{0.49\textwidth}
    \includegraphics[width=\textwidth]{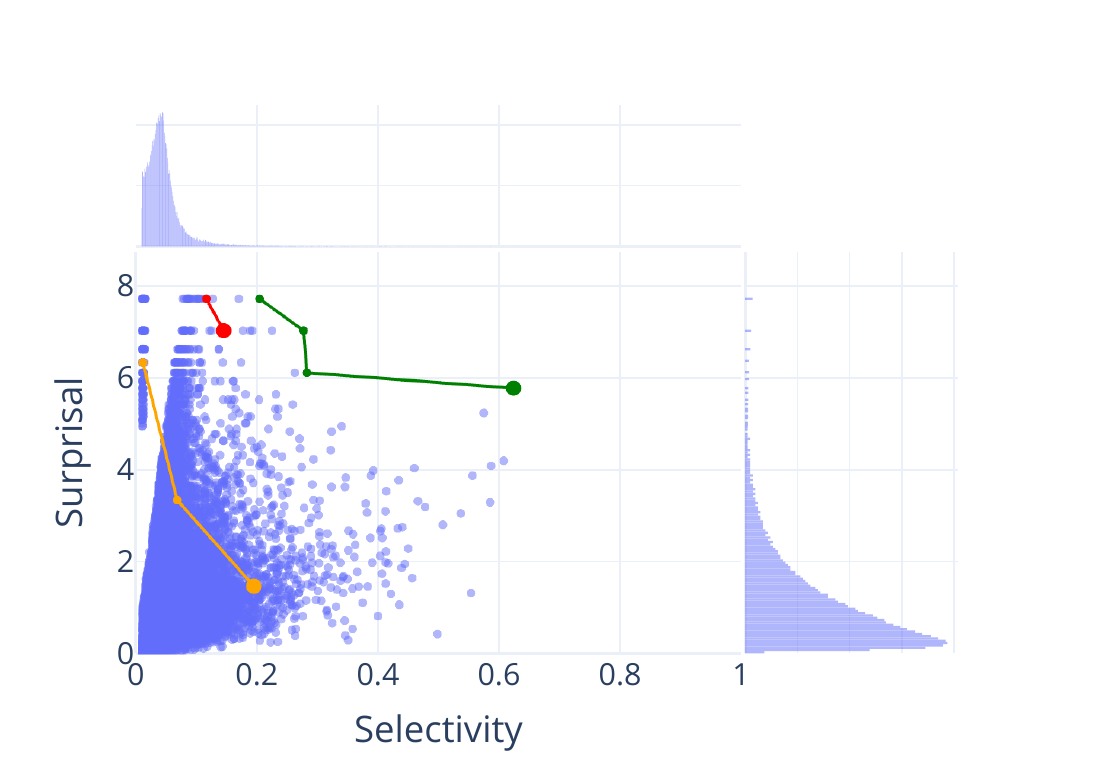}
    \caption{Hospitalization prediction.}
    \label{fig:task_3}
  \end{subfigure}
  \caption{Surprisal-selectivity Pareto fronts for datasets in which the surprisal and selectivity reached statistical significance ($p < 0.05$). The Pareto fronts obtained from our search procedure, sparse probes via SHAP values \citep{covert2021explaining,lundberg2017unified}, and optimal probing \citep{bertsimas2021sparse,gurnee2023finding} are shown in green, orange, and red, respectively. Larger markers indicate knee points. Both sparse probes via SHAP values and optimal probing are Pareto-dominated by our method. Values $< 0.01$ are omitted for improved readability.}
  \label{fig:task_1_3}
\end{figure}

\begin{figure}[h]
  \centering
  \begin{subfigure}{0.49\textwidth}
    \includegraphics[width=\textwidth]{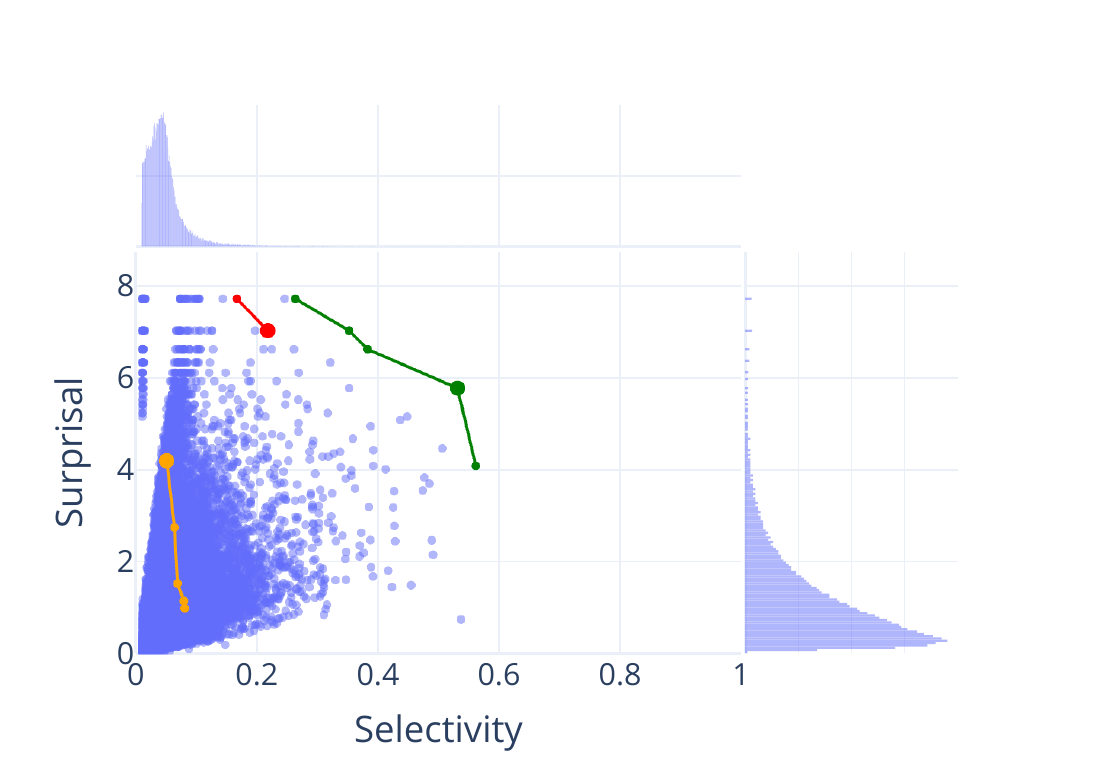}
    \caption{Inhospital mortality prediction.}
    \label{fig:task_0}
  \end{subfigure}
  \begin{subfigure}{0.49\textwidth}
    \includegraphics[width=\textwidth]{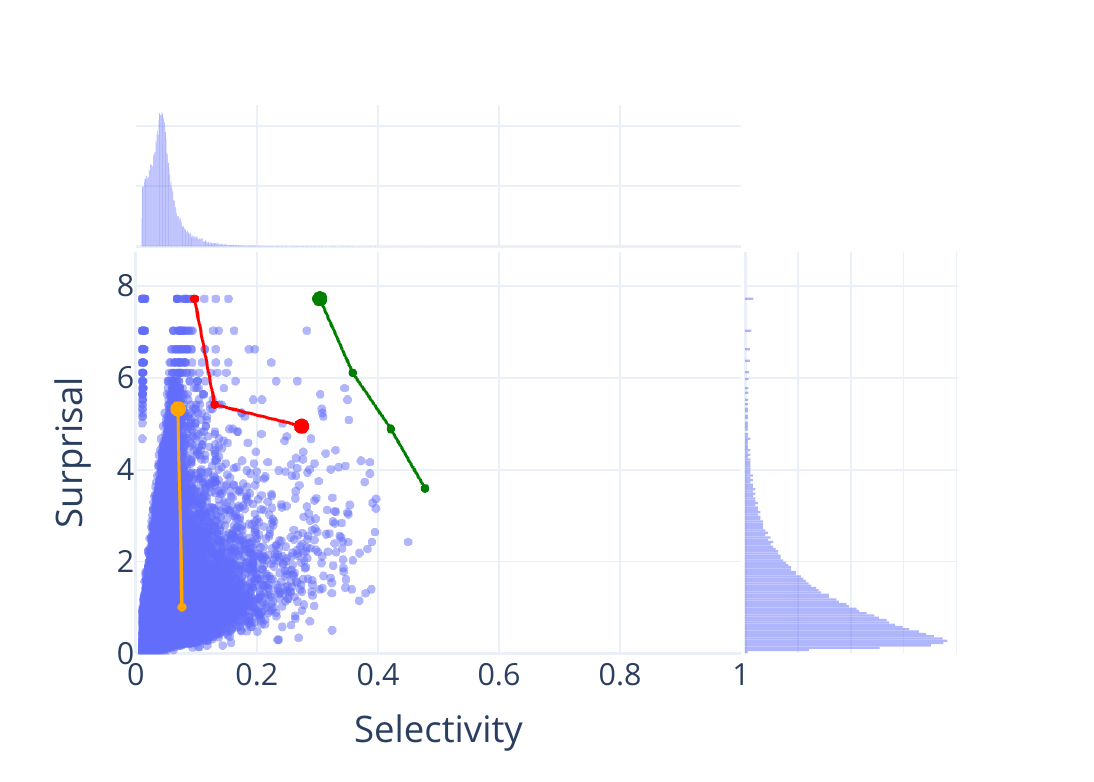}
    \caption{Critical outcome prediction.}
    \label{fig:task_2}
  \end{subfigure}
  \caption{Surprisal-selectivity Pareto fronts for datasets in which the surprisal and selectivity did not reach statistical significance ($p \geq 0.05$). The Pareto fronts obtained from our search procedure, sparse probes via SHAP values \citep{covert2021explaining,lundberg2017unified}, and optimal probing \citep{bertsimas2021sparse,gurnee2023finding} are shown in green, orange, and red, respectively. Larger markers indicate knee points. Both sparse probes via SHAP values and optimal probing are Pareto-dominated by our method. Values $< 0.01$ are omitted for improved readability.}
  \label{fig:task_0_2}
\end{figure}

\end{document}